\documentclass[11pt,letterpaper]{article}

\usepackage[utf8]{inputenc} 
\usepackage[T1]{fontenc}    
\usepackage{hyperref}       
\usepackage{url}            
\usepackage{booktabs}       
\usepackage{amsfonts}       
\usepackage{nicefrac}       
\usepackage{microtype}      
\usepackage[symbol]{footmisc}

\usepackage[margin=1in]{geometry} 
\usepackage{lmodern}
\usepackage[T1]{fontenc}
\usepackage[utf8]{inputenc}
\usepackage[tbtags]{amsmath}
\allowdisplaybreaks
\usepackage{amssymb,amsthm,mathtools, bbm}
\usepackage{physics}
\usepackage{hyperref}
\usepackage[svgnames]{xcolor} 
\usepackage[capitalise,nameinlink]{cleveref} 
\definecolor{links}{RGB}{11, 85, 255}
\definecolor{cites}{RGB}{0, 200, 0}
\definecolor{urls}{RGB}{255, 116, 0}
\hypersetup{colorlinks={true},linkcolor={links},citecolor=[named]{cites},urlcolor={urls}}
\usepackage[numbers,compress]{natbib} 
\usepackage{doi}
\usepackage{tikz} 
\usepackage{pgfplots}
\pgfplotsset{compat=1.14}
\usepgfplotslibrary{fillbetween}
\usepackage{hyphenat}
\usepackage[font=footnotesize]{caption}
\usepackage{subcaption}
\usepackage{appendix}
\usepackage{nicefrac}
\usepackage{tcolorbox}

\usepackage{thm-restate}

\usepackage{algorithm}
\usepackage[noend]{algpseudocode}

\usepackage{enumitem}
\newcommand{\cA}{\mathcal{A}}

\newcommand{\cD}{\mathcal{D}}
\newcommand{\cE}{\mathcal{E}}

\newcommand{\cH}{\mathcal{H}}

\newcommand{\cL}{\mathcal{L}}

\newcommand{\cS}{\mathcal{S}}

\newcommand{\cX}{\mathcal{X}}

\newcommand{\Top}{\textup{Top}}

\newcommand{\R}{\mathbb{R}}
\newcommand{\bbP}{\mathbb{P}}

\newcommand{\E}[2]{{\mathbb{E}}^{#1}\left[#2\right]}

\newcommand{\KL}[1]{{\cD_\textup{KL}}\left(#1\right)}
\renewcommand{\P}[2]{{\mathbb{P}}^{#1}\left[#2\right]}
\renewcommand{\exp}[1]{{\textup{exp}}\left(#1\right)}
\newcommand{\ind}[1]{\mathbbm{I}\left\{#1\right\}}

\newcommand{\alg}{\cA}

\newcommand{\VC}{\textmd{\textup{VC}}}
\newcommand{\LDim}{\textmd{\textup{LDim}}}

\newcommand{\Ber}[1]{\textmd{\textup{Ber}}\left(#1\right)}
\newcommand{\Hedge}{\textmd{\textup{Hedge}}}

\newcommand{\bl}{\boldsymbol{\ell}}

\newcommand{\bx}{\boldsymbol{x}}

\newcommand{\barmu}{\bar{\mu}}

\newcommand{\hatl}{\hat{\ell}}
\newcommand{\eps}{\varepsilon}

\theoremstyle{plain} 

\newtheorem{theorem}{Theorem}[section]
\newtheorem*{theorem*}{Theorem}
\newtheorem{lemma}[theorem]{Lemma}
\newtheorem{proposition}[theorem]{Proposition}
\newtheorem{corollary}[theorem]{Corollary}
\newtheorem{claim}[theorem]{Claim}

\newtheorem{observation}[theorem]{Observation}
\newtheorem{fact}[theorem]{Fact}

\newtheorem{example}[theorem]{Example}

\begin{document}

\title{Online Learning with Sublinear Best-Action Queries}
\date{}

\author{Matteo Russo\thanks{Sapienza University of Rome} \and Andrea Celli\thanks{Bocconi University} \and Riccardo Colini Baldeschi\thanks{Meta, UK} \and Federico Fusco\footnotemark[1] \and Daniel Haimovich\footnotemark[3] \and Dima Karamshuk\footnotemark[3] \and Stefano Leonardi\footnotemark[1] \and Niek Tax\footnotemark[3]}

\maketitle
\thispagestyle{empty}

\begin{abstract}
  In online learning, a decision maker repeatedly selects one of a set of actions, with the goal of minimizing the overall loss incurred. Following the recent line of research on algorithms endowed with additional predictive features, we revisit this problem by allowing the decision maker to acquire additional information on the actions to be selected. In particular, we study the power of \emph{best-action queries}, which reveal beforehand the identity of the best action at a given time step. In practice, predictive features may be expensive, so we allow the decision maker to issue at most $k$ such queries. We establish tight bounds on the performance any algorithm can achieve when given access to $k$ best-action queries for different types of feedback models. In particular, we prove that in the full feedback model,  $k$ queries are enough to achieve an optimal regret of $\Theta(\min\{\sqrt T, \nicefrac Tk\})$. This finding highlights the significant multiplicative advantage in the regret rate achievable with even a modest (sublinear) number $k \in \Omega(\sqrt{T})$ of queries. Additionally, we study the challenging setting in which the only available feedback is  obtained during the time steps corresponding to the $k$ best-action queries. There, we provide a tight regret rate of $\Theta(\min\{\nicefrac{T}{\sqrt k},\nicefrac{T^2}{k^2}\})$, which improves over the standard $\Theta(\nicefrac{T}{\sqrt k})$ regret rate for label efficient prediction for $k \in \Omega(T^{\nicefrac 23})$.
\end{abstract}

\section{Introduction}\label{sec:intro}
Online learning is a foundational problem in machine learning. In its simplest version, a decision maker repeatedly interacts with a fixed set of $n$ actions over a time horizon $T$. At each time, the decision maker needs to choose one of a set of actions; subsequently, it receives an action-dependent loss and observes some feedback.
These loss functions are generated by an omniscient (but oblivious) adversary and are only revealed on-the-go. The goal of the decision maker is to design a learning algorithm that achieves small {\em regret} with respect to the best fixed action in hindsight, i.e., the difference between the decision maker's loss and that of the fixed action. Several online learning algorithms have been developed, characterized by optimal instance-independent regret bound, depending on the feedback model \citep{CesaBL06,Slivkins19}.

Following the recent literature on algorithms with machine learning-based predictions (see, e.g., the survey by \citet{MitzenmacherV20}), we study the case where the learner is allowed to issue a limited number of \textit{best-action queries} to an oracle that reveals the identity of the best action for that step, so that the learner can choose it. This setting is motivated by scenarios in which obtaining accurate predictions on the optimal choice among numerous actions is possible but comes with significant costs and time constraints.
For instance, consider an online platform that continuously moderates posted content (e.g., Meta \citep{META1, META2} or Google \citep{GOOGLE1}), and the online learning problem it faces: posts are generated one after the other, and the platform’s task consists in deciding whether or not to flag the content as harmful. In this application, the platform may do so via (i) content moderation actions that are based on (expert) human reviews (that plays the role of \textit{best-action queries}), and (ii) automated content moderation decisions, i.e., decisions arising by employing an online learning algorithm. Due to budget or time constraints, access to human reviewing is a scarce resource, and the platform can only employ external reviewers at most $k$ times.

While the idea of incorporating hints or queries into online learning models has already been studied \citep[e.g., ][]{BhaskaraCKP21,BhaskaraGIKM23}, we are the first to study \textit{best-action queries}. \citet{BhaskaraCKP21} focus on online linear optimization with full feedback, with a query model which outputs vectors that are correlated with the actual losses. In the case of optimization over strongly convex domains, the regret bound improves from $\sqrt T$  to $\log T$, even for learners that receive hints for $O(\sqrt T)$ times. In an alternative model, \citet{BhaskaraGIKM23} studies comparison queries that allow the decision maker to know in advance, at each time, which among a small number of actions has the best loss. In this model, probing $2$ arms is sufficient to achieve time-independent regret bounds for online linear and convex optimization with full feedback, an in the stochastich multi-armed bandit problem. Our model differs from previous ones in two directions: (i) the online learner issues at most $k$ queries (differently from \citet{BhaskaraGIKM23}), and (ii) these queries are purely ordinal and the domain is not strongly convex\footnote{The model with $k$ actions can be captured by a linear function where each entry corresponds to the loss of a specific discrete action, and the continuous action space is the probability simplex over the $n$ discrete actions (which is not strongly convex).} (so that the  logarithmic bound in \citet{BhaskaraCKP21} does not apply). A more in-depth description of earlier work on correlated hints and ordinal queries in the context of online learning, as well as of the broader algorithms with predictions literature can be found in Appendix~\ref{sec:rel-work}.

\subsection{Our Model}\label{sec:model}

In our model, an online learner repeatedly interacts with $n$ actions over a time horizon $T$. At the beginning of each time $t \in [T]$\footnote{We adopt the notational convention that $[x]$ stands for the set of the first $x$ natural numbers.}, the learner chooses one of these actions $i_t$ and suffers a loss $\ell_t(i_t)$ generated by an (oblivious) adversary that may depend on both the action and the time; then, it observes some feedback. In this paper, we allow the learning algorithm $\cA$ to issue a \emph{best-action query}, for at most $k$ out of $T$ times. When the learner issues a query, an oracle reveals the \textit{identity} of the best action at that time, $i^*_t$, so that the learner can select it. The quality of a learning algorithm is measured via the regret: the difference between its performance and that of the best fixed action in hindsight. The regret of an algorithm $\cA$ against a sequence of losses $\bl \in [0,1]^{n \times T}$ reads
\begin{align*}
    R_T(\cA, \bl) = \sum_{t \in [T]} \E{}{\ell_t(i_t)} - \min_{i \in [n]} \sum_{t \in [T]} \ell_t(i),
\end{align*}
where the expectation runs over the (possibly) randomized decisions of the algorithm. We are interested in designing learning algorithms that perform well, i.e., suffer sublinear regret against all possible sequences of losses. For this reason, we denote with $R_T(\cA)$ (without the dependence on $\bl$) the worst-case regret of $\cA$:  $R_T(\cA) = \sup_{\bl} R_T(\cA,\bl).$ The minimax regret of a learning problem is then the regret achievable by the best algorithm. In our paper we pinpoint the exact minimax regret rates for the problems studied. 

For the sake of simplicity, we denote with $L_T(i)$ the total loss incurred by action $i$ over the whole time horizon: $L_T(i) =\sum_{t \in [T]} \ell_t(i)$. $L_T^{\min} = \sum_{t \in [T]} \ell_t(i^*_t)$ denotes instead the sum of the minimum loss actions at each time. Finally, $L_T(\cA_k) = \E{}{\sum_{t \in [T]} \ell_t(i_t)}$ denotes the expected total loss of an algorithm $\cA_k$ issuing at most $k$ best-action queries.

\subsection{Our Results}
\paragraph{Full feedback.} We start with the {\em full feedback} (a.k.a. prediction with experts) where the learner observes at each time the losses of all the actions after the action is chosen, i.e., {\em all} the loss vector $\bl_t = (\ell_t(1), \ell_t(2), \dots, \ell_t(n))$ after action $i_t$ is chosen. We obtain the following results:

\begin{itemize}
    \item We show that by combining the Hedge algorithm with $k$ queries issued uniformly at random (Algorithm \ref{alg:min-hedge}), we obtain a regret rate of $O(\min\{\sqrt{T}, \nicefrac Tk \})$, see Theorem~\ref{thm:adv2}. 
    \item 
    In Theorem~\ref{thm:lb}, we complement this positive result with a tight lower bound: we design a family of instances that forces every algorithm with $k$ queries to suffer the same regret. 
\end{itemize}

    It is surprising that issuing a sublinear number of queries, say $k = T^{\alpha}$ for $\alpha \in (\nicefrac 12, 1)$, yields a significant improvement on the regret, which becomes $T^{1-\alpha}.$ Note, the total of the losses incurred by any algorithm in $k$ rounds is at most $k$, which only affects the overall regret in an additive way; nevertheless, we prove that issuing $k$ queries has an impact on the regret that is \emph{multiplicative} in $k.$
    For instance, $T^{\nicefrac 23}$ queries are enough to decrease the regret to $T^{\nicefrac 13}$, which is well below the $\Theta(\sqrt{T})$ minimax regret for prediction with expert {\em without} queries \citep{CesaBL06}. 

\paragraph{Label efficient feedback.} We then proceed to study a partial feedback model inspired by the label efficient paradigm \citep{CesaBL06}. In the label-efficient prediction problem, the learner only observes the losses in a few selected times. With {\em label efficient feedback,} the learner only observes feedback during the $k$ times where the best-action queries are issued (but only after choosing the action). We obtain the following results:
\begin{itemize}
    \item We modify the (label-efficient) Hedge algorithm \citep{CesaBL06} to achieve a regret rate of $O(\min\{\nicefrac{T}{\sqrt{k}}, \nicefrac{T^2}{k^2}\})$, see Theorem~\ref{thm:adv3}, 
    \item In Theorem~\ref{thm:lb-le}, we show that it is not possible to improve on these rates: there exists instances where all algorithms suffer $\Omega(\nicefrac{T}{\sqrt{k}})$ regret if $k \in O(T^{\nicefrac 23})$ or $\Omega(\nicefrac{T^2}{k^2})$ if $k \in \Omega(T^{\nicefrac 23})$.
\end{itemize}

We observe that our algorithms improve (for $k \in \Omega(T^{\nicefrac 23})$) on the regret rate of label efficient prediction with $k$ steps of feedback, which is of order $\Theta(\nicefrac T{\sqrt{k}})$ \citep[Chapter 6.2 in][]{CesaBL06}. Also in this case, we observe the surprising multiplicative impact on the regret of the $k$ queries. For instance, $k = T^{\nicefrac 34}$ queries (which impact on a total loss of the same order but leave unaffected $\Theta(T)$ rounds) are enough to achieve $O(\sqrt{T})$ regret. 

\paragraph{Stochastic i.i.d. setting.} In Section~\ref{sec:stoch-mab}, we derive (up to polylogarithmic factors) the above results in the stochastic setting, but with different algorithms. Namely, in full feedback, we show how Follow-The-Leader \citep{AuerCBFS03} achieves regret $\Theta(\nicefrac{T}{k})$ when $k \geq \Omega(\sqrt{T})$, and $\tilde \Theta(\sqrt{T})$ otherwise. With label efficient feedback, Explore-Then-Commit \citep{PerchetRCS15} achieves $\tilde \Theta(\nicefrac{T^2}{k^2})$ when $k \geq \Omega(T^{\nicefrac{2}{3}})$, and $\tilde \Theta(\nicefrac{T}{\sqrt{k}})$ otherwise.\footnote{We use $\tilde \Theta(x)$ to hide polylogarithmic terms in $x$.}

\subsection{Technical Challenges}

    Issuing a best-action query has the effect of inducing a possibly \emph{negative} instantaneous regret. In fact, the benchmark that the algorithm is comparing against at time $t$ is the loss of the best fixed action over the whole time horizon, which may naturally be larger, at time $t$, than the best action $i_t^*$ in that specific time. Overall, the magnitude of the total ``negative'' regret that is generated by $k$ queries is at most $O(k)$, which is typically sublinear in $T$, and affects {\em linearly} on the overall regret. We prove that this sublinear number of queries has a (surprising) multiplicative effect on the overall regret bound. 

Our analysis reveals that any algorithm employing a \textit{uniform querying} strategy can be characterized by a total loss decomposed into two terms: a $\nicefrac kT$ fraction of the best dynamic loss, representing the sum of the smallest losses at each time, and the total loss of the same algorithm operating without queries, scaled by a factor of $1-\nicefrac kT$. Additionally, within our proof, we conduct a more refined analysis of the Hedge algorithm in the \textit{no-querying} setting. Specifically, we can express the regret as a function of the difference between the best dynamic loss and the best-fixed cumulative loss in hindsight. This formulation demonstrates a noteworthy enhancement in regret rates compared to the conventional bound of $O(\sqrt{T})$ when the discrepancy between the best dynamic loss and the best-fixed cumulative loss in hindsight is relatively small.

Concerning lower bounds, their construction is more challenging than their ``no queries'' counterpart. In fact, we need to design hard instances against \emph{any} query-augmented learning algorithm, and provide tight bounds in both $k$ and $T.$ The lower bounds we construct are stochastic, this implies that the minimax regret we find are tight, even in the stochastic model, where the losses are drawn i.i.d. from a fixed but unknown distribution.


\section{Related Work}\label{sec:rel-work}

\paragraph{Correlated hints.} A first model that is close to ours has been introduced by \citet{DekelFHJ17}, which studies Online Linear Optimization when the learner has access to vectors correlated with the actual losses. Surprisingly, this type of hint guarantees an exponential improvement in the regret bound, which is logarithmic in the time horizon when the optimization domain is strongly convex. Subsequently, \citet{BhaskaraC0P20} generalize such results in the presence of {\em imperfect hints}, while \citet{BhaskaraGIKM23} prove that $\Omega(\sqrt{T})$ hints are enough to achieve the logarithmic regret bound. While these works share significant similarities with ours, we stress that their results are not applicable to the standard prediction with experts model, given the non-strong convexity of the probability simplex over the actions.

\paragraph{Queries/Ordinal hints.} \citet{BhaskaraGIKM23} studies online learning algorithms augmented with ordinal queries of the following type: a query takes in input a small subset of the actions and receives in output the identity of the best one. With this twist, it is possible to bring the regret down to $O(1)$ by observing only $2$ experts losses in advance at each time. Although our best-action query is stronger (as it compares {\em all} the actions), our learner is constrained in the number of times it can issue such queries.

\paragraph{Algorithms with predictions.} Other works that explore the interplay between hints and feedback are \citep{BhaskaraC0P21,ShiLJ22,ChengZJ23,BhaskaraM23,BhaskaraC0P23}. More broadly, our work follows the literature on enhancing  performance through external information (predictions).  This has been extensively applied to a variety of online problems to model partial information about the input sequence that can be fruitfully exploited if accurate for improving the performance of the algorithms. Examples include sorting \citep{BaiC23}, frequency estimation \citep{AamandCNSV23}, various online problems \citep{PurohitSK18, GollapudiP19} such as metrical task systems \citep{AntoniadisCEPS23,AntoniadisCEPS23-b}, graph coloring \citep{AntoniadisBM23}, caching \citep{LykourisV21}, scheduling \citep{LattanziLMV20, JiangPS22}, and several others.

\section{Hedge with $k$ Best-Action Queries}\label{sec:ub}

In this section, we propose two algorithms that address online learning with full and label efficient feedback. They are built by combining the Hedge algorithm \citep[Chapter 2 in][]{CesaBL06} to uniform queries. We start by presenting some known properties of Hedge (in full feedback) that are crucial key for the main results of this section.


\begin{restatable}{lemma}{lempotentialhedge}\label{lem:potential-hedge}
    Consider the Hedge algorithm $\Hedge_\eta(\tilde \bl)$ run on loss sequence $\tilde \bl \in [0,U]^{n \times T}$ with learning rate $\eta< \nicefrac{1}{U}$. Then, for all action $i \in [n]$, it holds that,
    \[
         \tilde L_T(\Hedge_\eta(\tilde \bl)) \leq \frac{1}{1-U\eta} \cdot \left(\tilde L_T(i) + \frac{\log n}{\eta}\right),
    \]
    where $\tilde L_T(\Hedge_\eta(\tilde \bl)) = \sum_{t \in [T], i \in [n]} p_t(i) \cdot \tilde \ell_t(i)$ is the expected cumulative loss of $\Hedge_\eta(\tilde \bl)$ and $\tilde L_T(i)$ is the cumulative loss of action $i$.
\end{restatable}


\begin{proof}
    We have that, by definition,
    \begin{align*}
        W_{T+1} \geq w_{T+1}(i) 
            w_t(i) \cdot \exp{-\eta\tilde\ell_t(i)} = \prod_{t \in [T]} \exp{-\eta\tilde\ell_t(i)} = \exp{-\eta \tilde L_T(i)}.
    \end{align*}
    We also know that 
    \begin{align*}
        W_{t+1} &\leq W_t \cdot \left(1 - \eta\sum_{i \in [n]} p_t(i)\cdot\tilde\ell_t(i) + \eta^2\sum_{i \in [n]} p_t(i)\cdot\tilde\ell^2_t(i)\right).
    \end{align*}
    Thus,
    \begin{align*}
        W_{T+1} &\leq n \cdot \prod_{t \in [T]} \left(1 - \eta\sum_{i \in [n]} p_t(i) \cdot \tilde\ell_t(i) + \eta^2\sum_{i \in [n]} p_t(i) \cdot \tilde\ell^2_t(i)\right),
    \end{align*}
    which combined with the earlier bound and taking logarithms of both sides, gives
    \begin{align*}
            -\eta\tilde L_T(i) &\leq \log n+\sum_{t \in [T]} \log \left( 1 - \eta\sum_{i \in [n]} p_t(i)\cdot \tilde\ell_t(i) +\eta^2\sum_{i \in [n]} p_t(i) \cdot \tilde\ell^2_t(i) \right)\\
            &\leq \log n - \eta \sum_{t \in [T], i \in [n]} p_t(i) \cdot \tilde\ell_t(i) + \eta^2\sum_{t \in [T], i \in [n]} p_t(i)\cdot \tilde\ell^2_t(i)\\
            &\leq \log n - (\eta - U\eta^2) \sum_{t \in [T], i \in [n]} p_t(i)\cdot \tilde\ell_t(i).
        \end{align*}
    The second inequality above follows from $\log(1+z) \leq z$ for $z \in \R$, and the third by observing that $\tilde\ell^2_t(i) \leq U\tilde\ell_t(i)$, since $\tilde\ell_t(i) \in [0,U]$. The lemma follows by rearranging the terms above.
\end{proof}


\subsection{Full Feedback: An $O(\frac{T\log n}{k})$ Regret Bound}\label{sec:ub-adv}

\begin{theorem}\label{thm:adv2}
    Consider the problem of online learning with full feedback and $k$ best-action queries, then there exists an algorithm $\cA_k$ that guarantees
    \[
        R_T(\cA_k) \leq \min\left\{\sqrt{T\log n}, \frac{T \log n}{k}\right\}.
    \]
\end{theorem}
     We prove that  Algorithm~\ref{alg:min-hedge} exhibits the desired regret guarantees. We have the following theorem. In particular, as we illustrate next, this algorithm performs uniform querying, i.e., they choose a uniformly random subset $Q \subseteq [T]$ of size $k$ where to issue best-action queries. In the case of uniform queries, a useful simplification can be made.

\begin{observation}\label{obs:adv-reg-mab}
    Let $\cA_0$ be an algorithm with no querying power, and consider $\cA_k$, obtained from $\cA_0$ by equipping it with $k$ uniformly random queries across the time horizon $T$. Similarly, let $i_t^0$ and $i_t$ be the actions selected by $\cA_0$ and $\cA_k$ at time $t$. Then, for all adversarial sequences $\bl \in [0,1]^{n \times T}$ of action losses, 
    \[
        \E{}{\ell_t(i_t)} = \left(1 - \frac{k}{T}\right) \cdot \E{}{\ell_t(i_t^0)} + \frac{k}{T} \cdot \E{}{\ell_t(i_t^*)},
    \]
    for all $t \in [T]$, and thus
    \begin{align}\label{eq:alg-rand-query}
        L_T(\cA_k) = \left(1 - \frac{k}{T}\right) \cdot L_T(\cA_0) + \frac{k}{T} \cdot L^{\min}_T.
    \end{align}
\end{observation}

\begin{algorithm}
    \caption{Hedge with Best-Action Queries}
    \label{alg:min-hedge}
    \begin{algorithmic}[1]
        \State \textbf{Input:} Sequence of losses $\ell_t(i)$ and query budget $k \in [T]$
        \State Sample $k$ out of $T$ time steps uniformly at random and denote this random set by $Q$
        \State Set $\eta = \sqrt{\frac{\log n}{T}}$ when $k \leq \sqrt{T}$, otherwise $\eta = \frac{k}{T}$
        \State Initialize $w_1(i) = 1$ for all $i \in [n]$
        \For{$t \in [T]$}
            \If{$t \in Q$}
                \State Observe $i^*_t = \arg\min_{i \in [n]} \ell_t(i)$
                \State Select action $i^*_t$
            \Else
                \State Let $W_t = \sum_{i \in [n]} w_t(i)$
                \State Select action $i$ with probability $p_t(i) = \frac{w_t(i)}{W_t}$
            \EndIf
            \State Observe $\ell_t(i)$ for all $i \in [n]$
            \State Update $w_{t+1}(i) = w_t(i) \cdot \exp{-\eta(\ell_t(i)-\ell_t(i^*_t))}$ for all $i \in [n]$
        \EndFor
    \end{algorithmic}
\end{algorithm}

\begin{proof}[Proof of Theorem~\ref{thm:adv2}]
    Let us first note that Algorithm~\ref{alg:min-hedge} without queries is an instantiation $\Hedge_\eta(\tilde \bl)$ applied to losses $\tilde \ell_t(i) = \ell_t(i)-\ell_t(i^*_t) \in [0,1]$. Let this algorithm be denoted as $\cA_0$. Then, applying Lemma~\ref{lem:potential-hedge} and expanding the terms, we obtain
    \[
        L_T(\cA_0) \leq  \frac{L_T(i)}{1-\eta} +\frac{\log n}{\eta(1-\eta)} - \frac{\eta L_T^{\min}}{1-\eta}. 
    \]
    Let $\cA_k$ be Algorithm~\ref{alg:min-hedge} with $k$ best-action queries. By Observation~\ref{obs:adv-reg-mab}, specifically \eqref{eq:alg-rand-query}, it holds that
    \begin{align*}
        &L_T(\cA_k) \leq \frac{(1-\nicefrac{k}{T})L_T(i)}{1-\eta} + \frac{(1-\nicefrac{k}{T})\log n}{\eta(1-\eta)} - \frac{\eta(1-\nicefrac{k}{T}) L_T^{\min}}{1-\eta} + \frac{k}{T} \cdot L_T^{\min} \\
        \Longleftrightarrow~ &R_T(\cA_k, i) \leq \frac{(1-\nicefrac{k}{T})\log n}{\eta(1-\eta)} + \frac{T\eta-k}{T(1-\eta)} (L_T(i) - L_T^{\min}) \leq \min\left\{\sqrt{T\log n}, \frac{T \log n}{k}\right\},
    \end{align*}
     where the last inequality holds by setting $\eta = \max\left\{\sqrt{\nicefrac{\log n}{T}}, \nicefrac{k}{T}\right\}$.
\end{proof}

\subsection{Label Efficient Feedback: An $O(\frac{T^2\log n}{k^2})$ Regret Bound} \label{app:hedgelim}

We extend Algorithm~\ref{alg:min-hedge} to a setting where feedback is given only during querying time steps, with the only difference that the update rule is performed just after the querying time steps and nowhere else across the time horizon. We prove the following theorem:
\begin{theorem}\label{thm:adv3}
    For all adversarial sequences $\bl \in [0,1]^{n \times T}$ of action losses and for all $k \geq \sqrt{\frac{T \log T}{2}} - 1$, in the label efficient query model, there exists an algorithm $\cA_k$ that guarantees
    \[
        R_T(\cA_k) \leq 2\cdot \min\left\{T\sqrt{\frac{2\log n}{k}}, \frac{T^2 \log n}{k^2}\right\}.
    \]
\end{theorem}

\paragraph{Algorithm description.} We first describe the algorithm $\cA_k$ we use: Let $X_t \sim \Ber{\nicefrac{\hat k}{T}}$ be a Bernoulli random variable, for some $\hat k \leq k$ to be specified later. $\cA_k$ issues a best action query if, at time step $t$, $X_t = 1$ and unless the query budget is exhausted. Once the query budget is exhausted, the algorithm stops querying. Otherwise, it performs the usual update rule on losses $\hat \ell_{t+1}(i) = \frac{T}{\hat k} \cdot (\ell_t(i) - \ell_t(i^*_t)) \cdot \ind{X_t = 1}$. The algorithm then simply selects action $I_t=i^*_t$ if $X_t = 1$ and action $I_t=i$ with probability $p_t(i)$ if $X_t = 0$. Moreover, we denote by $X_{\leq t} = (X_1, \ldots, X_t), I_{\leq t} = (I_1, \ldots, I_t)$.

For the sake of the analysis, we introduce another algorithm $\cA_k^\prime$, which is the same as algorithm $\cA_k$ with the only (but crucial) difference that it issues a query if and only if $X_t = 1$, regardless of whether or not query budget is exhausted. We thus bound the regret of $\cA_k$ in terms of the regret of $\cA^\prime_k$.

\begin{lemma}\label{lem:le-expectation}
    For all adversarial sequences $\bl \in [0,1]^{n \times T}$ of action losses, in the label efficient query model, algorithm $\cA^\prime_k$ guarantees
    \[
        R_T(\cA^\prime_k) \leq \min\left\{T\sqrt{\frac{2\log n}{\hat k}}, \frac{T^2 \log n}{\hat k^2}\right\}.
    \]
\end{lemma}
\begin{proof}
    For algorithm $\cA_k^\prime$, we have that its counterpart without queries, $\cA_0^\prime$, is an instantiation $\Hedge_\eta(\tilde \bl)$, with $\tilde \bl = \hat \bl$. Thus, by Lemma~\ref{lem:potential-hedge}, we obtain
    \begin{align}
        \hat L_T(\cA^\prime_k) = \sum_{t \in [T], i \in [n]} p_t(i) \cdot \hat \ell_t(i) \leq \frac{1}{1-\frac{T}{\hat k}\eta} \cdot \left(\hat L_T(i) + \frac{\log n}{\eta}\right), \label{eq:le-eq}
    \end{align}
    as long as $\eta < \nicefrac{\hat k}{T}$. We now recognize that, since $\E{}{\hat \ell_t(i) \mid X_{\leq t-1}, I_{\leq t-1}} = \ell_t(i) - \ell_t(i^*_t)$, then
    \begin{align*}
        \sum_{i \in [n]}\E{}{ p_t(i) \cdot \hat \ell_t(i) \bigg\vert X_{\leq t-1}, I_{\leq t-1}} = \sum_{i \in [n]} p_t(i) \cdot (\ell_t(i) - \ell_t(i^*_t)).
    \end{align*}
    Therefore, by the tower property of expectation applied around \eqref{eq:le-eq}, we have
    \begin{align*}
        L_T(\cA^\prime_0) &= \sum_{i \in [n]}\E{}{ p_t(i) \cdot \hat \ell_t(i) } = \sum_{i \in [n]} p_t(i) \cdot (\ell_t(i) - \ell_t(i^*_t)) \\
        &\leq \frac{1}{1-\frac{T}{\hat k}\eta} \cdot \left(L_T(i) - L_T^{\min} + \frac{\log n}{\eta}\right) = \frac{\hat k}{\hat k-T\eta} \cdot \left(L_T(i) + \frac{\log n}{\eta}\right) - \frac{T\eta}{\hat k-T\eta} L_T^{\min}.
    \end{align*}
    By Observation~\ref{obs:adv-reg-mab}, we get
    \begin{align*}
        L_T(\cA^\prime_k) &\leq \left(1 - \frac{k}{T}\right) \cdot \frac{\hat k}{\hat k-T\eta} \cdot \left(L_T(i) + \frac{\log n}{\eta}\right) - \frac{(T-k)\eta}{\hat k-T\eta} L_T^{\min} + \frac{k}{T}L_T^{\min},
    \end{align*}
    which means that the regret is upper bounded by
    \begin{align*}
        R_T(\cA^\prime_k, i) \leq \frac{\hat k\log n}{\eta(\hat k-T\eta)} + \frac{T^2\eta - k\hat k}{T(\hat k-T\eta)}(L_T(i) - L_T^{\min}) 
        \leq \min\left\{T\sqrt{\frac{2\log n}{\hat k}}, \frac{T^2 \log n}{\hat k^2}\right\},
    \end{align*}
    where the last inequality holds by setting $\eta = \max\left\{\frac{1}{T}\sqrt{\frac{\hat k\log n}{2}}, \frac{k\hat k}{\sqrt{2}T^2}\right\}$, and then by noticing, in the latter case, that $\hat k-\frac{k\hat k}{\sqrt{2}T} \geq \sqrt{2}k$.
\end{proof}
With this lemma, we are ready to prove Theorem~\ref{thm:adv3}, where we bound the regret of algorithm $\cA_k$.
\begin{proof}[Proof of Theorem~\ref{thm:adv3}]
    We consider event $\cE = \{|Q| \leq k\}$, so that, slightly abusing notation, we write the regret of algorithm $\cA_k$ as
    \begin{align*}
        R_T(\cA_k) &= \E{}{R_T(\cA_k) \mid \cE} \cdot \P{}{\cE} + \E{}{R_T(\cA_k) \mid \bar \cE} \cdot \P{}{\bar \cE} \leq \E{}{R_T(\cA_k) \mid \cE} + T \cdot \P{}{\bar \cE}.
    \end{align*}
    To upper bound the second summand above, we have
    \begin{align}
        T \cdot\P{}{\bar \cE} \leq T \cdot \exp{-\frac{2(k+1-\hat k)^2}{T}} \leq T \cdot \frac{1}{T} = 1,\label{eq:help}
    \end{align}
    by Hoeffding's inequality applied on the binomial random variable $|Q|$ with expectation $\hat k$, and as long as $\hat k \geq k - \sqrt{\frac{T\log T}{2}} + 1$.
    
    For what concerns the first summand, we recognize that under event $\cE$, $\cA_k$ and $\cA^\prime_k$ are exactly the same algorithm. Thus, it holds that $\E{}{R_T(\cA_k, i) \mid \cE} = \E{}{R_T(\cA^\prime_k, i) \mid \cE}$. Moreover, if $\E{}{R_T(\cA_k) \mid \bar \cE} \geq 0$, and since $\P{}{\cE} \geq 1 - \nicefrac{1}{T}$, then
    \[
        \E{}{R_T(\cA^\prime_k, i) \mid \cE} \leq \frac{T}{T-1} \cdot R_T(\cA_k^\prime, i) \leq \frac{T}{T-1} \cdot \min\left\{T\sqrt{\frac{2\log n}{\hat k}}, \frac{T^2 \log n}{\hat k^2}\right\},
    \]    
    by Lemma~\ref{lem:le-expectation}. If, instead, $\E{}{R_T(\cA_k) \mid \bar \cE} < 0$, we also know that, by an identical derivation to \eqref{eq:help}, $\E{}{R_T(\cA_k) \mid \bar \cE} \cdot \P{}{\bar \cE} \geq -1$. Therefore,
    \[
        \E{}{R_T(\cA^\prime_k, i) \mid \cE} \leq \frac{T}{T-1} \cdot (R_T(\cA_k^\prime, i) + 1) \leq \frac{T}{T-1} \cdot \left(\min\left\{T\sqrt{\frac{2\log n}{\hat k}}, \frac{T^2 \log n}{\hat k^2}\right\} + 1\right),
    \]
    again by Lemma~\ref{lem:le-expectation}. Overall, we obtain
    \[R_T(\cA_k) \leq 2\cdot \min\left\{T\sqrt{\frac{2\log n}{k}}, \frac{T^2 \log n}{k^2}\right\}. \qedhere\]
\end{proof}
\section{Lower Bounds}\label{sec:lower-bounds}

    In this section, we construct two of  randomized instances of the learning problem, which induce a tight lower bound on the minimax regret rates for both feedback models. We define the random variables $Z_t$ as the feedback observed by the algorithm at the end of time $t$. In the full feedback model, $Z_t = \ell_t$, while in the label efficient setting, $Z_t = \ell_t$ only if a query is issued at time $t$ (and $Z_t$ is an empty $n$-dimensional vector). Furthermore, we denote with $Z_{\leq t}$ the array containing the feedback $Z_1, Z_2, \dots, Z_t$ until time $t$.

    We start by describing the two stochastic instances, which are characterized by two distributions over two losses. As a notational convention, we denote the losses with $\ell_t$, and introduce two probability measures $\bbP^+, \bbP^-$ (and their corresponding expectations $\mathbb{E}^+, \mathbb{E}^-$). Let $\varepsilon, q \in [0,1]$ be two parameters we set later (with $\varepsilon \le q$), we have that the losses of the $n = 2$ actions are distributed as follows: 
    \[
        (\ell_t(1), \ell_t(2)) = \begin{cases}
            (1,1) &\text{ w.p. } \frac{1}{2} \text{ under both $\bbP^+$ and $ \bbP^-$}\\
            (0,0) &\text{ w.p. } \frac{1}{2} - 2q \text{ under both $\bbP^+$ and $ \bbP^-$}\\
            (0,1) &\text{ w.p. } q + \varepsilon \text{ under $\bbP^+$ and w.p. }q - \varepsilon \text{ under $\bbP^-$} \\
            (1,0) &\text{ w.p. } q - \varepsilon \text{ under $\bbP^+$ and w.p. }q + \varepsilon \text{ under $\bbP^-$}
        \end{cases}.
    \]


    We now introduce and prove a general Lemma on the expected regret $R_T^{\pm}(\cA_k)$ suffered by any deterministic algorithm $\cA_k$ which issues at most $k$ queries, against the i.i.d. sequence of valuations drawn according to $\bbP^{\pm}$. Since we want a result that holds for both feedback models, we introduce the random set $F$ which contains the times where $\cA_k$ actually observes the losses; note, $N_F = |F|$ is equal to $T$ in full feedback and to the number queries issued in the partial information model.


\begin{lemma}\label{lem:lb}
    For any determinstic algorithm $\cA_k$ which issues at most $k$ best-action queries, we have:
    \begin{align*}
        R^+_T(\cA_k) + R^-_T(\cA_k) &\geq \exp{-\frac{5\varepsilon^2}{q}\E{+}{N_F}} \cdot \frac{(T-k)\varepsilon}{2} -2(q-\varepsilon)k.
    \end{align*}
\end{lemma}

\begin{proof}
    The best action $i^*$ under $\bbP^+$, respectively $\bbP^-$ is the first, respectively the second, one, with an expected loss of 
    \begin{equation}
    \label{eq:loss_istar}
        \E{\pm}{\ell_t(i^*)} = \tfrac 12 + q - \eps.
    \end{equation}

    On the other hand, the best realized action $i^*_t$ yields an expected loss of 
    \begin{equation}
        \label{eq:loss_istart}
        \E{\pm}{\ell_t(i^*_t)} = \E{\pm}{\min\{\ell_t(1), \ell_t(2)\}} = -(q-\varepsilon).
    \end{equation}
    Moreover, if the algorithm chooses a suboptimal action $i_t \neq i^*$, its expected instantaneous regret is:
    \begin{equation}
        \label{eq:error}
        \E{+}{\ell_t(2)} - \E{+}{\ell_t(i^*)} = \E{-}{\ell_t(1)} - \E{-}{\ell_t(i^*)} = 2 \eps.
    \end{equation}
        
    Let now $N_+$, respectively $N_-$, be the random variable that counts the number of times that $\cA_k$ selects action $1$, respectively $2$, in times that are not in $F$ (i.e., where the choice of $1$ is not due to a query). Combining \eqref{eq:loss_istar},\eqref{eq:loss_istart}, and \eqref{eq:error}, we have the following:
    \begin{align*}
        R_T^{\pm}(\cA_k) &\ge \E{\pm}{ \sum_{t \notin F} (\ell_t(i_t) - \ell_t(i^*)) + \sum_{t \in F} (\ell_t(i^*_t) - \ell_t(i^*))} \\
        &= 2 \eps \cdot \E{\pm}{N_{\mp}} - (q-\eps)\E{\pm}{N_F}.
    \end{align*}
    Since $N_F \le k$, and $N_+ + N_- \le T - k$, we have the following bound on the regret:
    %
    \begin{align*}
        R^{+}_T(\cA_k) &\geq \P{+}{N_+ \leq \frac{T-k}{2}} \cdot {(T-k)\varepsilon} - (q-\varepsilon)k\\
        R^{-}_T(\cA_k) &\geq \P{-}{N_+ > \frac{T-k}{2}} \cdot {(T-k)\varepsilon} - (q-\varepsilon)k.
    \end{align*}
    Summing the above two expressions, we obtain
    \begin{align}
        R^{+}_T(\cA_k) + R^{-}_T(\cA_k) &\geq \left(\P{+}{N_+ \leq \frac{T-k}{2}} + \P{-}{N_+ > \frac{T-k}{2}}\right) \cdot (T-k)\varepsilon -2(q-\varepsilon)k. \label{eq:reg-intermediate-k}
    \end{align}

    At this point, we apply the Bretagnolle-Huber Inequality \citep[Theorem 14.2 in][]{LattimoreS20} to bound the first term on the right-hand side of the above inequality:
    \begin{align*}
        \P{+}{N_+ \leq \frac{T-k}{2}} + \P{-}{N_+ > \frac{T-k}{2}} &\geq \frac{1}{2} \cdot \exp{-\KL{\bbP^+_{Z_{\leq T}}, \bbP^-_{Z_{\leq T}}}},
    \end{align*}
    where $\bbP^{\pm}_{Z \le T}$ is the push-forward measure on all the possible sequences of feedback observed by $\cA_k$ under $\bbP^{\pm}$. The lemma is concluded by combining the above inequality with the following claim, and plugging it into \eqref{eq:reg-intermediate-k}.
    \begin{claim}
        It holds that $
        \KL{\bbP^+_{Z_{\leq T}}, \bbP^-_{Z_{\leq T}}} \leq \frac{5\varepsilon^2}{q}\E{+}{N_F}.$
    \end{claim}
    \begin{proof}
    Let us observe that, once we fix the feedback history until time $t-1$, i.e., fix a realization of the feedback $Z_{\leq t-1}$, 
    we have that
    \begin{align*}
        \KL{\bbP^+_{Z_t \mid Z_{\leq t-1}}, \bbP^-_{Z_t \mid Z_{\leq t-1}}} = \ind{t \in F} \cdot \KL{\bbP^+_{Z_t \mid t \in F}, \bbP^-_{Z_t \mid t \in F}},
    \end{align*}
    where $\bbP^+_{Z_t \mid Z_{\leq t-1}}$ (respectively $\bbP^-_{Z_t \mid Z_{\leq t-1}}$) is the push-forward measure over $\{0,1\}^2$ when losses are drawn according to $\bbP^+$ (respectively $\bbP^-$), conditioning on the previous observations. The equality above holds because (i) algorithm $\cA_k$ observes feedback if and only $t \in F$ (by definition), and (ii) $\cA_k$ is deterministic and whether or not $t \in F$ may only depend on the past. We now upper bound the KL-divergence term above: 
    \begin{align*}
        \KL{\bbP^+_{Z_t \mid Z_{\leq t-1}}, \bbP^-_{Z_t \mid Z_{\leq t-1}}} &= \left(q + \varepsilon\right) \cdot \log\left(1 + \frac{2\varepsilon}{q-\varepsilon}\right)+\left(q - \varepsilon\right) \cdot \log\left(1 - \frac{2\varepsilon}{q+\varepsilon}\right)\\
        &\leq (q+\varepsilon) \cdot \frac{2\varepsilon}{q-\varepsilon} - (q-\varepsilon) \cdot\frac{2\varepsilon}{q+\varepsilon} = \frac{4\varepsilon^2q}{q^2-\varepsilon^2} \leq \frac{5\varepsilon^2}{q},
    \end{align*}
    where the first inequality follows from $\log(1+z) \leq z$ for all $z \in \R$, and the last holds as long as we choose $\varepsilon < \frac{q}{\sqrt{5}}$. To complete our derivation, we express the overall KL divergence exploiting the tower property of conditional expectation: 
    \begin{align*}
        \KL{\bbP^+_{Z_{\leq T}}, \bbP^-_{Z_{\leq T}}} &= \sum_{t \in [T]} \E{+}{\KL{\bbP^+_{Z_t \mid Z_{\leq t-1}}, \bbP^-_{Z_t \mid Z_{\leq t-1}}}} \leq \frac{5\varepsilon^2}{q}\E{+}{N_F},
    \end{align*}
     where expectation is taken over all possible feedback realizations $Z_{\leq t-1}$, and the last inequality follows by earlier derivations. 
    \end{proof}
    This concludes the proof of the lemma.
\end{proof}

We now show how to use the above lemma to derive the lower bounds. We start with full feedback.

\begin{theorem}\label{thm:lb}
    In the full feedback query model, for all $k \in [T]$, we have the following lower bounds: 
    \begin{itemize}
        \item For any algorithm $\cA_k$ that has access to $k < c_0 \sqrt T$ queries, it holds that $R_T(\cA_k) \ge c_0\frac{\sqrt{T}}{4}$.
        \item For any algorithm $\cA_k$ that has access to $k \ge c_0 \sqrt T$ queries, it holds that $R_T(\cA_k) \ge c_1 \frac{T}{k}$.
    \end{itemize}
    Where $ c_0 = \nicefrac{1}{(e^8\sqrt{5})}$ and $c_1 = \nicefrac{1}{(320 e^2)}$ are universal constants.
\end{theorem}

\begin{proof}
    In full feedback, the algorithm always observes the losses, so that the feedback variable $Z_t= \ell_t$ and $N_F = T$. We prove the Theorem via Yao's minimax Theorem: we prove that any deterministic algorithm $\cA_k$ fails against the random instance composed as follows: with probability $\nicefrac 12$ the losses are drawn i.i.d. according to $\bbP^+$, otherwise, they are drawn i.i.d. according to $\bbP^-$. We can then apply Lemma~\ref{lem:lb} and obtain that the expected regret of $\cA_k$ against such mixture is equal to
   \begin{equation}
       \label{eq:full_final}
       \E{}{R_T(\cA_k)} = \frac 12 (R^+_T(\cA_k) + R^-_T(\cA_k)) \geq \exp{-\frac{5\varepsilon^2}{q}T} \cdot \frac{(T-k)\varepsilon}{2} -2(q-\varepsilon)k.
   \end{equation}
   Now, if $k < c_0 \sqrt T$, we set $\varepsilon = \frac{2}{\sqrt{5T}}$ and $q = \frac{1}{4}$ in \eqref{eq:full_final} to get
    \[
        \E{}{R_T(\cA_k)} = \frac 12 (R^+_T(\cA_k) + R^-_T(\cA_k)) \geq \frac{\sqrt{T}}{2 e^8\sqrt{5}} - \frac{\sqrt{T}}{4e^8\sqrt{5}} \geq \frac{\sqrt{T}}{4e^8\sqrt{5}},
    \]
    Otherwise, if $k \geq c_0 \sqrt T$, then consider the following choice of the parameters: $\varepsilon = \frac{1}{40e k} + \frac{4e-1}{40e T}$ and $q = 5\varepsilon^2 T = \frac{{T}}{{320e^2k^2}} + \frac{{4e-1}}{{160e^2k}} + \frac{{(4e-1)^2}}{{320e^2T}}$. Plugging these parameters in \eqref{eq:full_final}, we get:
    \begin{align*}
        \E{}{R_T(\cA_k)} = \frac 12 (R^+_T(\cA_k) + R^-_T(\cA_k)) &\geq \frac{T\varepsilon}{4e} - 5\varepsilon^2kT + \left(1-\frac{1}{4e}\right)\cdot \varepsilon k \\
        &= \frac{{T}}{{320e^2k}} + \frac{{4e-1}}{{160e^2}} + \frac{{(4e-2)^2k}}{{320e^2T}} \geq \frac{T}{320e^2 k}.\qedhere
    \end{align*}
\end{proof}

As similar analysis can be carried over for the label efficient setting.
\begin{theorem}\label{thm:lb-le}
    In the label efficient feedback model, for all $k \in [T]$, we have the following lower bounds: 
    \begin{itemize}
        \item For any algorithm $\cA_k$ that has access to $k < c_0 \tfrac{T}{\sqrt k}$ queries, it holds that $R_T(\cA_k) \ge c_0 \frac{T}{4 \sqrt k}$.
        \item For any algorithm $\cA_k$ that has access to $k \ge c_0 \tfrac{T}{\sqrt k}$ queries, it holds that $R_T(\cA_k) \ge c_1 \frac{T^2}{k^2}$.
    \end{itemize}
    Where $ c_0 = \nicefrac{1}{(e^8\sqrt{5})}$ and $c_1 = \nicefrac{1}{(320 e^2)}$ are universal constants.
\end{theorem}

\begin{proof}
    In the label efficient feedback model, $Z_t$ is meaningful only for times in $F$. We then prove the result by Yao's minimax principle, by showing that any determinisitic algorithm $\cA_k$ which issues at most $k$ queries suffers the desired regret against the instance that uniformly chooses between $\bbP^+$ and $\bbP^-$. We can apply Lemma~\ref{lem:lb} (nothing that $N_F \le k$) to get:
   \begin{align*}
       \E{}{R_T(\cA_k)} = \frac 12 (R^+_T(\cA_k) + R^-_T(\cA_k))
       &\geq \exp{-\frac{5\varepsilon^2}{q}k} \cdot \frac{(T-k)\varepsilon}{4} - (q-\varepsilon)k.
   \end{align*}
   Once again, we have two cases. If $k \geq \frac{T}{e^8\sqrt{5k}}$, we choose $\varepsilon = \frac{2}{\sqrt{5k}}$ and $q = \frac{1}{4}$ to get
    \[
        \E{}{R_T(\cA_k)} = \frac 12 (R^+_T(\cA_k) + R^-_T(\cA_k))\geq \frac{T}{2e^8\sqrt{5k}} - \frac{T}{4e^8\sqrt{5k}} \geq \frac{T}{4e^8\sqrt{5k}}.
    \]
    Otherwise, if $k \geq \frac{T}{e^8\sqrt{5k}}$, and we choose $\varepsilon = \frac{T}{40e k^2} + \frac{4e-1}{40e k}$ and $q = 5\varepsilon^2 T = \frac{{T^2}}{{320e^2k^3}} + \frac{{(4e-1)T}}{{160e^2k^2}} + \frac{{(4e-1)^2}}{{320e^2k}}$, to obtain
    \begin{align*}
        \E{}{R_T(\cA_k)} = \frac 12 (R^+_T(\cA_k) + R^-_T(\cA_k)) &\geq \frac{T\varepsilon}{4e} - 5\varepsilon^2k^2 + \left(1-\frac{1}{4e}\right)\cdot \varepsilon k \\
        &= \frac{{T^2}}{{320e^2k^2}}  + \frac{{T(4e - 1)}}{{160e^2k}} + \frac{{(4e - 1)^2}}{{320e^2}} \geq \frac{T^2}{320e^2 k^2}. \qedhere
    \end{align*}
\end{proof}

\section{Best-Action Queries in the Stochastic i.i.d. Setting}\label{sec:stoch-mab} 

In this section, we show how the results provided in Section ~\ref{sec:ub} can be obtained in the stochastic i.i.d. setting (defined next) using the Follow-The-Leader algorithm, albeit a suboptimal dependence in $n$.

In the stochastic i.i.d. setting, each action is associated with a fixed but unknown distribution $D_i$ supported in $[0,1]$. Action $i$'s loss at time step $t$,  $\ell_{t}(i)$, is drawn independently from distribution $D_i$. We denote with $\mu(i)$ the expected loss of action $i$, and with $i^* = \arg\min_{i \in [n]} \mu(i)$ be the action of lowest expected loss. We measure the performance of a learning algorithm $\cA$, by considering its regret with respect to the best action distribution:
\begin{align*}
    R_T(\cA) = \sum_{t \in [T]} \E{}{\ell_t(i_t) - \ell_t(i^*)}.
\end{align*}
We denote by $\Delta_i = \mu(i) - \mu(i^*)$ be the gap between the expected loss of the best action and that of action $i$, and by $\Psi_i = \E{}{\abs{\ell(i)-\ell({i^*})}}$ the expected absolute value of such gap (note, we omit the dependence on time as the losses are drawn i.i.d. across time). Whenever the learner issues a query, then the identity of the action $i^*_t$ with best {\em realized} loss  is revealed, so that the learner gets, in expectation, $ \E{}{\ell_{t}(i^*_t)} = \E{}{\min_{i \in [n]} \ell_{t}(i)}$. 

\subsection{Useful Facts and Observations}

The following facts and simple results are particularly useful for analyzing algorithms in the stochastic generation model.

\begin{fact}[Bernstein's Inequality]\label{fct:bernstein}
    Let $Y_1, \ldots, Y_m$ be independent random variables such that $\sum_{\tau=1}^m \E{}{Y_\tau} = \mu$ and $\P{}{|Y_\tau| \leq c} = 1$, for $c > 0$ and all $\tau \in [m]$. Then, for $\gamma > 0$, we have 
    \[
        \P{}{\sum_{\tau=1}^{m} Y_\tau - \mu \geq \gamma} \leq \exp{-\frac{3\gamma^2}{6\sigma^2 + 2c\gamma}},
    \]
    where $\sigma^2 = \sum_{\tau = 1}^m \E{}{Y^2_\tau} - \E{}{Y_\tau}^2$. Moreover, when $Y_\tau$'s are also identically distributed, we have $\sigma^2 = m(\E{}{Y^2_\tau} - \E{}{Y_\tau}^2) \leq m\E{}{|Y_\tau|} =: m\Psi$. Thus,
    \[
        \P{}{\sum_{\tau=1}^{m} Y_\tau - \mu \geq \gamma} \leq \exp{-\frac{3\gamma^2}{6m\Psi + 2c\gamma}}.
    \]
\end{fact}

\begin{proposition}\label{prop:stoch-reg-mab}
    Let $T, n \in \mathbb{N}$ be the time horizon and number of experts respectively, with $T \geq n-1$. For all distributions $D_i$ of actions losses, and any algorithm $\cA$, its regret is
    \begin{align}\label{eq:regret-mab}
        R_T(\cA_k) &= \sum_{t=k+1}^T \Delta_{i_t} - \frac{k}{2} \cdot (\Psi-\Delta),
    \end{align}
    where $\Psi := \E{}{\abs{\min_{i\neq i^*} \ell(i) - \ell(i^*)}}$ and $\Delta := \E{}{\min_{i\neq i^*} \ell(i)-\ell(i^*)}$.
\end{proposition}
\begin{proof}
    We first observe that, by the i.i.d. assumption, we can always make the algorithm query in the first $k$ time steps, without losing generality. Therefore, as $Q = \{t \mid t \leq k\}$, the algorithm suffers $\E{}{\min_{i \in [n]} \ell_t(i)}$ in the first $k$ time steps, which, by independence, gives
    \[
        \sum_{t \leq k} \E{}{\min_{i \in [n]} \ell_t(i)} = k\E{}{\min_{i \in [n]} \ell(i)}.
    \]
    We also have the following useful observation that allows us to rewrite the maximum in a convenient form. Namely,
    \begin{align}\label{eq:rewrite}
        \min_{i \in [n]} \ell(i) = \frac{\ell(i^*)+\underset{i \neq i^*}{\min}~\ell(i) - \abs{\underset{i \neq i^*}{\min} ~\ell(i) - \ell(i^*)}}{2}.
    \end{align}
    The claim follows by the definitions of $\Psi$ and $\Delta$.
\end{proof}

Thus, in the stochastic case, the goal of a regret-minimizing algorithm is to minimize the first summand above since the second is characteristic of the instance at hand.

\begin{observation}\label{obs:psidelta}
    Let $T, n \in \mathbb{N}$ be the time horizon and number of experts respectively, with $T \geq n-1$. For all distributions $D_i$ of actions losses, $\Psi_i - \Delta_i \leq \Psi - \Delta$.
\end{observation}
\begin{proof}
    Since $\E{}{\min_{j \in [n]} \ell(j)} \leq \E{}{\min\{\ell(i^*), \ell(i)\}}$ for all $i \in [n]$, we have
    \begin{align*}
        \Delta - \Psi &= 2 \cdot \E{}{\min_{j \in [n]} \ell(j) - \ell(i^*)}\leq 2 \cdot \E{}{\min\{\ell(i), \ell(i^*)\} - \ell(i^*)} =  \Delta_i - \Psi_i,
    \end{align*}
    where the last equality also follows from Equation \eqref{eq:rewrite} applied to actions $i, i^*$ only.
\end{proof}

\subsection{Label Efficient Feedback:Explore-Then-Commit with $\Omega(T^{\nicefrac 23})$ Best-Action Queries}

We consider the label efficient feedback case first, where feedback is given only during querying time steps. The techniques used in this section are key for the full feedback case, and easier to illustrate.

We provide a simple variation of the Explore-Then-Commit (ETC) algorithm \cite{PerchetRCS15} that, in the first $k \geq \Omega(T^{\nicefrac 23})$ time steps, is given free access to the identity of the best action before committing to the action to select. 
\begin{algorithm}
    \caption{Explore-Then-Commit with Best-Action Queries}
    \label{alg:etc}
    \begin{algorithmic}
        \State \textbf{Input:} Sequence of losses $\ell_t(i) \sim D_i$ and $k \in [T]$ query budget
        \For{$t \in [T]$}
            \If{$t \leq k$}
                \State Observe $i^*_t = \arg\min_{i \in [n]} \ell_t(i)$
                \State Select action $i^*_t$
                \State Observe $\ell_t(i)$ for all $i \in [n]$
            \Else
                \State Select action $j = \arg\min_{i \in [n]} \barmu_k(i)$
            \EndIf
        \EndFor
    \end{algorithmic}
\end{algorithm}
We have the following theorem:
\begin{restatable}{theorem}{thmetc}\label{thm:etc}
    Let $T, n \in \mathbb{N}$ be the time horizon and number of actions respectively, with $T \geq n-1$. For all distributions $D_i$ of actions losses, Algorithm~\ref{alg:etc} with query access $k \in [\nicefrac{4T}{9}]$ guarantees
    \[
        R_T(\cA_k) \leq \min\left\{3T\sqrt{\frac{\log 2nT}{2k}}, \frac{2nT^2\ln T}{k^2}\right\}.
    \]
\end{restatable}

We split the proof of this theorem in two lemmas that immediately imply it, one for $k < T^{\nicefrac 23}$ and one otherwise.

\begin{lemma}
    If $k < T^{\nicefrac 23}$, then $R_T(\cA_k) \leq 3T\sqrt{\frac{\log 2nT}{2k}}$.
\end{lemma}
\begin{proof}
    For all actions $i$, we define the clean event as
    \[
        \cE_i = \{|\overline \mu_k(i) - \mu(i)| < \varepsilon\},
    \]
    where $\varepsilon=\sqrt{\frac{\log 2nT}{2k}}$. Similarly, let $\cE$ be the intersection of all the $\cE_i$'s for $i \in [n]$. By Hoeffding's inequality \citep[Equation (5.8) in Chapter 5 of][]{LattimoreS20}, we have the following:
    \[
        \P{}{\bar\cE} = \P{}{\cup_{i \in [n]} \bar\cE_i} \le \sum_{i \in [n]}\P{}{\bar\cE_i} \le 2n \cdot \exp{-{2k \varepsilon^2}} \le \tfrac 1T.
    \]
    Consider now the expected instantaneous regret suffered by ETC at time $t+1$. We have two cases: Either the clean event holds, so that the instantaneous regret is at most $2\varepsilon_t$, or it does not, in which case the instantaneous regret is at most $1$. By the law of total probability, we have the following:
    \[
        \E{}{\ell_{t+1}(i_t) - \ell_{t+1}(i)} \le \P{}{\bar\cE_t} + \E{}{\ell_{t+1}(i_t) - \ell_{t+1}(i)|\cE_t} \le \tfrac 1T + 2\varepsilon_t \le 3 \varepsilon_t.
    \]
    Summing the expected instantaneous regrets for all $t$ yields the desired bound of
    \[R_T(\cA_k) \leq 3T\sqrt{\frac{\log 2nT}{2k}},\]
    as we use the lower bound $\Psi-\Delta \geq 0$ to say that in the first $k$ times the regret the algorithm suffers is at most $0$.    
\end{proof}

\begin{lemma}
    If $T^{\nicefrac 23} \leq k \leq \nicefrac{4T}{9}$, then $R_T(\cA_k) \leq \frac{2nT^2\ln T}{k^2}$.
\end{lemma}

\begin{proof}
We start by showing the statement for 2 actions, and then generalize to $n$.

\paragraph{The case of 2 actions.} Let us recall that, by Proposition~\ref{prop:stoch-reg-mab}, we have
\begin{align*}
    R_T(\cA_k) = \abs{\{t \geq k+1 \mid i_t = 2\}} \cdot \Delta - \frac{k}{2} \cdot (\Psi - \Delta),
\end{align*}
where we have assumed that $\mu_1 \leq \mu_2$, without loss of generality. We suffer positive regret in the last $T-k$ time steps when we select action $2$ over $1$, which happens if and only if $\barmu_k(1) \leq \barmu_k(2)$. Hence, the first summand of the above regret expression simply becomes $\Delta(T-k) \cdot \P{}{\barmu_k(2) \leq \barmu_k(1)}$. 

Note that $$\P{}{\barmu_k(2) \leq \barmu_k(1)} = \P{}{\sum_{t=1}^k \ell_t(1)-\ell_t(2) \geq 0},$$ so if we define $g_t = \ell_t(1) - \ell_t(2)$, whose expected value is $\E{}{g_t} = -\Delta$ and $\P{}{|g_t| \leq 1} = 1$, then, by Fact~\ref{fct:bernstein-weak},
\begin{align*}
    \P{}{\sum_{t=1}^k g_t \geq 0} &= \P{}{\sum_{t=1}^k g_t + k\Delta \geq k\Delta} \\
    &\leq \exp{-\frac{3k^2\Delta^2}{6k\Psi + 2k\Delta}} \\
    &\leq \exp{-\frac{k\Delta^2}{2\Psi + \Delta}}.
\end{align*}
We distinguish two cases, namely $\frac{\Delta^2}{2\Psi + \Delta} \leq \frac{\ln T}{k}$ and $\frac{\Delta^2}{2\Psi + \Delta} > \frac{\ln T}{k}$. In the latter case, we get that 
\begin{align*}
    R_T(\cA_k) &= \Delta(T-k)\cdot \P{}{\barmu_k(2) \leq \barmu_k(1)} - \frac{k}{2} \cdot (\Psi - \Delta) \\
    &\leq \Delta T\cdot \exp{-\frac{k\Delta^2}{2\Psi + \Delta}} - \frac{k}{2} \cdot (\Psi - \Delta) \\
    &\leq \Delta T \cdot \exp{-\ln T} = \Delta \leq 1.
\end{align*}
Otherwise, we obtain that $\Psi \geq \frac{\Delta}{2} \cdot \left(\frac{k\Delta}{\ln T} - 1 \right)$, and so
\begin{align*}
    R_T(\cA_k) &= \Delta(T-k)\cdot \P{}{\barmu_k(2) \leq \barmu_k(1)} - \frac{k}{2}\cdot(\Psi - \Delta) \\
    &\leq \Delta T - \frac{k\Delta}{4} \cdot \left(\frac{k\Delta}{\ln T} - 3 \right)\\
    &\leq \left(\frac{T^2}{k^2} + \frac{3T}{4k}-\frac{27}{16}\right) \cdot \ln T \\
    &\leq \frac{2T^2\ln T}{k^2}.
\end{align*}
The first inequality holds since the maximizing $\Delta =\frac{4T-3k}{2k^2} \cdot \ln T$, which is consistent with $0 \leq \Delta \leq \Psi \leq 1$ since $T^{\nicefrac 23} \leq k \leq \nicefrac{4T}{9}$.

\paragraph{Generalizing to $n$ actions.} We again assume that the first is the best action, and apply Proposition~\ref{prop:stoch-reg-mab}, to obtain
\begin{align*}
    R_T(\cA_k) &=\sum_{\substack{t=k+1\\j\neq 1}}^T\E{}{\ind{\barmu_k(j) \geq \max_{i \neq j} \barmu_{k}(i)} (\ell_t(j)-\ell_t(1))} - \frac{k}{2}\cdot(\Psi-\Delta)\\
    &= (T-k) \cdot \sum_{j \neq 1} \Delta_j \cdot \P{}{\barmu_k(j) \geq \max_{i \neq j} \barmu_k(i)} - \frac{k}{2}\cdot(\Psi-\Delta)\\
    &\leq T \cdot \sum_{j \neq 1} \Delta_j \cdot \P{}{\barmu_k(j) \geq \barmu_k(1)}   - \frac{k}{2}\cdot(\Psi-\Delta),
\end{align*}
which holds by the independence of realizations across time steps. Let us define $g_t(j) = \ell_t(1)-\ell_t(j)$, so that, $\Psi_j = \E{}{|g_t(j)|}$, $\E{}{g_t(j)} = -\Delta_j$ and $\P{}{|g_t(j)| \leq 1} = 1$. Then, by Fact~\ref{fct:bernstein-weak}, we obtain
\begin{align*}
    \P{}{\barmu_k(j) \geq \barmu_k(1)} &= \P{}{\sum_{t=1}^k g_t(j) \geq 0} \leq \exp{-\frac{k\Delta_j^2}{2\Psi_j + \Delta_j}}.
\end{align*}
Recall that, by Observation~\ref{obs:psidelta}, $\Psi - \Delta \geq \Psi_j - \Delta_j$. Therefore, let us sum and subtract to the regret upper bound the sum $\frac{k}{2n} \cdot \sum_{j \neq 1} (\Psi_j - \Delta_j)$, and get
\begin{align*}
    R_T(\cA_k) &\leq\sum_{j \neq 1} \left(T\Delta_j \cdot \exp{-\frac{k\Delta_j^2}{2\Psi_j + \Delta_j}} - \frac{k}{2n} \cdot (\Psi_j - \Delta_j)\right)  \\
    &\quad + \frac{k}{2} \cdot \left(\frac{1}{n} \cdot \sum_{j \neq 1} (\Psi_j - \Delta_j) - (\Psi-\Delta)\right)\\
    &\leq \sum_{j \neq 1} \left(T\Delta_j \cdot \exp{-\frac{k\Delta_j^2}{2\Psi_j + \Delta_j}} - \frac{k}{2n} \cdot (\Psi_j - \Delta_j)\right).
\end{align*}
For each $j \neq 1$, an identical derivation to the one in the case of two actions would yield the same regret rate, with $k/n$ in place of $k$. Thus, overall, we obtain
\[
    R_T(\cA_k) \leq \frac{2nT^2\ln T}{k^2},
\]
which concludes the proof.
\end{proof}

\subsection{Full Feedback: Follow-The-Leader with $\Omega\left(\sqrt{T}\right)$ Best-Action Queries}

We provide an algorithm that, equipped with best-action queries for $k$ time steps and with full feedback, achieves regret $O(\nicefrac Tk)$ for $k \geq \Omega(\sqrt{T})$, and $O(\sqrt{T})$ otherwise. This means that, with the addition of feedback, fewer queries are needed to switch to a much smaller regret rate. 

The idea is to use the Follow-The-Leader paradigm \cite{AuerCBFS03}. That is, we select the best action in the first $k$ time steps since queries give us its identity for free, and successively, we select the action maximizing the empirical average so far, $\barmu_{t-1}(i) = \frac{1}{t-1}\cdot\sum_{\tau=1}^{t-1} \ell_\tau(i)$. 
\begin{algorithm}
    \caption{Follow-The-Leader with Best-Action Queries}
    \label{alg:ftl}
    \begin{algorithmic}
        \State \textbf{Input:} Sequence of losses $\ell_t(i) \sim D_i$ and $k \in [T]$ query budget
        \For{$t \in [T]$}
            \If{$t \leq k$}
                \State Observe $i^*_t = \arg\min_{i \in [n]} \ell_t(i)$
                \State Select action $i^*_t$
                \State Observe $\ell_t(i)$ for all $i \in [n]$
            \Else
                \State Select action $j = \arg\min_{i \in [n]} \barmu_{t-1}(i)$
            \EndIf
        \EndFor
    \end{algorithmic}
\end{algorithm}

We have the following guarantee:
\begin{restatable}{theorem}{thmftl}\label{thm:ftl}
    Let $T, n \in \mathbb{N}$ be the time horizon and number of actions respectively, with $T \geq n-1$. For all distributions $D_i$ of actions losses, Algorithm~\ref{alg:ftl} with query access $k \geq 2\sqrt{T}$ guarantees
    \[
        R_T(\cA_k) \leq \min\left\{3\sqrt{2T\log 2nT}, \frac{5nT}{k}\right\}.
    \]
\end{restatable}

For our purposes, only a weaker version of Fact~\ref{fct:bernstein} is needed:
\begin{fact}\label{fct:bernstein-weak}
    Let $Y_1, \ldots, Y_m$ be i.i.d. random variables such that $\E{}{Y_\tau} = -\Delta_Y$, $\E{}{|Y_\tau|} = \Psi_Y$, and $\P{}{|Y_\tau| \leq 1} = 1$, for all $\tau \in [m]$. Then, it holds that
    \[
        \P{}{\sum_{\tau=1}^{m} Y_\tau \geq 0} \leq \exp{-\frac{m\Delta_Y^2}{2\Psi_Y + \Delta_Y}}.
    \]
\end{fact}

We split the proof of this theorem in two lemmas that immediately imply it, one for $k < 2\sqrt{T}$ and one otherwise. Next, we subsume the notation of Theorem~\ref{thm:ftl}.

\begin{lemma}
    If $k < 2\sqrt{T}$, then $R_T(\cA_k) \leq 3\sqrt{2T\log 2nT}$.
\end{lemma}
\begin{proof}
    For all times $t$ and actions $i$, we define the clean event as
    \[
        \cE_{i,t} = \{|\overline \mu_t(i) - \mu(i)| < \varepsilon_t\},
    \]
    where $\varepsilon_t = \sqrt{\frac{\log 2nT}{2t}}$. Similarly, let $\cE_t$ be the intersection of all the $\cE_{i,t}$'s for $i \in [n]$. By Hoeffding's inequality \citep[Equation (5.8) in Chapter 5 of][]{LattimoreS20}, we have the following:
    \[
        \P{}{\bar\cE_t} = \P{}{\cup_{i \in [n]} \bar\cE_{i,t}} \le \sum_{i \in [n]}\P{}{\bar\cE_{i,t}} \le 2n \cdot \exp{-{2t \varepsilon^2_t}} \le \tfrac 1T.
    \]
    Consider now the expected instantaneous regret suffered by FTL at time $t+1$. We have two cases: Either the clean event holds, so that the instantaneous regret is at most $2\varepsilon_t$, or it does not, in which case the instantaneous regret is at most $1$. By the law of total probability, we have the following:
    \[
        \E{}{\ell_{t+1}(i_t) - \ell_{t+1}(i)} \le \P{}{\bar\cE_t} + \E{}{\ell_{t+1}(i_t) - \ell_{t+1}(i)|\cE_t} \le \tfrac 1T + 2\varepsilon_t \le 3 \varepsilon_t.
    \]
    Summing the expected instantaneous regrets for all $t$ yields the desired bound of
    \[R_T(\cA_k) \leq 3\sqrt{2T\log 2nT},\]
    as we use the lower bound $\Psi-\Delta \geq 0$ to say that in the first $k$ times the regret the algorithm suffers is at most $0$.
\end{proof}

\begin{lemma}
    If $k \geq 2\sqrt{T}$, then $R_T(\cA_k) \leq \frac{5nT}{k}$.
\end{lemma}
\begin{proof}
    We start by showing the statement for 2 actions, and then generalize to $n$.
    \paragraph{The case of 2 actions.} We again assume that $\mu_1 \leq \mu_2$ without loss of generality. FTL's regret, thus, reads
    \begin{align}\label{eq:ftl}
        R_T(\cA_k) &= \E{}{\sum_{t=k+1}^T (\ell_t(2)-\ell_t(1)) \cdot \ind{\barmu_{t-1}(2) \leq \barmu_{t-1}(1)}} - \frac{k}{2}\cdot(\Psi-\Delta) \nonumber\\
        &= \Delta \cdot \sum_{t=k+1}^T \P{}{\barmu_{t-1}(2) \leq \barmu_{t-1}(1)} - \frac{k}{2}\cdot(\Psi-\Delta) \nonumber\\
        &\leq \Delta \cdot \sum_{t=k+1}^T \exp{-\frac{t\Delta^2}{2\Psi + \Delta}} - \frac{k}{2}\cdot(\Psi-\Delta),
    \end{align}
where the second equality follows from the independence of step $t$ from all the preceding steps, and the inequality by Fact~\ref{fct:bernstein-weak}. We distinguish three cases, namely $\frac{\Delta^2}{2\Psi + \Delta} \leq \frac{1}{T}$, $\frac{1}{T} < \frac{\Delta^2}{2\Psi + \Delta} \leq \frac{\ln T}{k}$, and $\frac{\Delta^2}{2\Psi + \Delta} > \frac{\ln T}{k}$. In the first case, we have that $\Psi \geq \frac{\Delta}{2} \cdot (T\Delta-1)$ and so,
\begin{align*}
    R_T(\cA_k) &\leq \Delta T - \frac{k\Delta}{4} \cdot (T\Delta-3) \\
    &\leq \frac{T}{k} + \frac{3}{2} + \frac{27k}{64T} \\
    &\leq \frac{2T}{k},
\end{align*}
where the second inequality follows since the $\Delta$ maximizing the expression is $\Delta = \frac{2}{k} + \frac{3}{4T}$. This implies that $\Psi \geq \frac{T}{2k^2} - \frac{3}{32T} + \frac{1}{2k}$, which is consistent with $0 \leq \Delta \leq \Psi \leq 1$ since $k \geq 2\sqrt{T}$.

In the last case, $\frac{\Delta^2}{2\Psi + \Delta} > \frac{\ln T}{k}$, we get
\begin{align*}
    R_T(\cA_k) &= \Delta(T-k)\cdot \P{}{\barmu_{t-1}(2) \leq \barmu_{t-1}(1)} - \frac{k}{2} \cdot (\Psi - \Delta) \\
    &\leq \Delta T\cdot \exp{-\frac{t\Delta^2}{2\Psi + \Delta}} - \frac{k}{2} \cdot (\Psi - \Delta) \\
    &\leq \Delta T \cdot \exp{-\ln T} = \Delta\\
    & \leq 1.
\end{align*}
We are left to show the case where $\frac{1}{T} < \frac{\Delta^2}{2\Psi + \Delta} \leq \frac{\ln T}{k}$. To this end, let us observe that, for all $T \geq 2$, it holds that
\[
    \sum_{t=m+1}^{T} \exp{-\frac{t}{r}} \leq r.
\]
for all $1 \leq m, r \leq T$. This is the case because
\begin{align*}
    \sum_{t=m+1}^{T} \exp{-\frac{t}{r}} = \frac{e^{-k/r} - e^{-T/r}}{e^{1/r}-1} \leq r e^{-k/r} \leq r,
\end{align*}
since $1+z \leq e^z$ for all $z \in \R$. Therefore,
\begin{align*}
    R_T(\cA_k) &\leq 2\Delta \cdot \frac{2\Psi + \Delta}{\Delta^2} - \frac{k}{2}(\Psi - \Delta) = \Psi \cdot \left(\frac{4}{\Delta} - \frac{k}{2}\right) + 2 + \frac{k\Delta}{2}.
\end{align*}
We distinguish two subcases: On the one hand, assume that $\Delta > \frac{8}{k}$, then $\frac{4}{\Delta} - \frac{k}{2} < 0$. Since $\Psi \geq \frac{k\Delta^2}{2\ln T} - \frac{\Delta}{2}$, we have
\begin{align*}
    R_T(\cA_k) &\leq \left(\frac{k\Delta^2}{2\ln T} - \frac{\Delta}{2}\right) \cdot \left(\frac{4}{\Delta} - \frac{k}{2}\right) + 2 + \frac{k\Delta}{2} \\
    &= \frac{2k\Delta}{\ln T} - \frac{k\Delta^2}{4\ln T} + \frac{3k}{4} \\
    &\leq \frac{3\ln T}{8} + 4 + \frac{1}{\ln T}\left(8-\frac{4}{k}\right) - \frac{1}{k} \cdot \left(\frac{\ln T}{16} - 1\right) \\
    &\leq \frac{\ln T}{2},
\end{align*}
since the expression is maximized for $\Delta = \frac{8 + \ln T}{2k} > \frac{8}{k}$, as long as $T > e^{8}$. This implies that $\Psi \geq \frac{8}{k\ln T} - \frac{\ln T}{8k}$, which is consistent with $0 \leq \Delta \leq \Psi \leq 1$. On the other hand, $\Delta \leq \frac{8}{k}$, and thus $\frac{4}{\Delta} - \frac{k}{2} \geq 0$. We use that $\Psi \leq \frac{T\Delta^2 - \Delta}{2}$, and get
\begin{align*}
    R_T(\cA_k) &\leq \frac{T\Delta^2 - \Delta}{2} \cdot \left(\frac{4}{\Delta} - \frac{k}{2}\right) + 2 + \frac{k\Delta}{2} \\
    &= 2T\Delta - \frac{kT\Delta^2}{4} + \frac{3k\Delta}{4}\\
    &\leq \frac{4T}{k} + \frac{9k}{16T} \\
    &\leq \frac{5T}{k},
\end{align*}
since the expression is maximized for $\Delta = \frac{4}{k} + \frac{3}{2T} < \frac{8}{k}$. This implies that $\Psi \leq \frac{8T}{k^2} + \frac{4}{k} + \frac{3}{8T}$, which is consistent with $0 \leq \Delta \leq \Psi \leq 1$ since $k \geq 2\sqrt{T}$.

\paragraph{Generalizing to $n$ actions.} The generalization is very similar to the one in the proof of ~\ref{thm:etc}. Indeed, the regret equals
\begin{align*}
    R_T(\cA_k) &= \sum_{\substack{t=k+1\\j\neq 1}}^T\E{}{\ind{\barmu_{t-1}(j) \leq \min_{i \neq j} \barmu_{t-1}(i)} (\ell_t(j)-\ell_t(1))} - \frac{k}{2}\cdot(\Psi-\Delta)\\
    &= \sum_{t=k+1}^T\sum_{j \neq 1} \Delta_j \cdot \P{}{\barmu_{t-1}(j) \leq \min_{i \neq j} \barmu_{t-1}(i)} - \frac{k}{2}\cdot(\Psi-\Delta)\\
    &\leq \sum_{t=k+1}^T\sum_{j \neq 1} \Delta_j \cdot \P{}{\barmu_{t-1}(j) \leq \barmu_{t-1}(1)}   - \frac{k}{2}\cdot(\Psi-\Delta)\\
    &\leq \sum_{j \neq 1} \Delta_j \cdot \sum_{t=k+1}^T \exp{-\frac{t\Delta_j^2}{2\Psi_j+\Delta_j}}   - \frac{k}{2}\cdot(\Psi-\Delta),
\end{align*}
where the last inequality follows by Fact~\ref{fct:bernstein-weak}. Identically to the proof of Theorem~\ref{thm:etc}, we sum and subtract the term $\frac{k}{2n} \cdot \sum_{j \neq 1} (\Psi_j - \Delta_j)$, and get 
\begin{align*}
    R_T(\cA_k) &\leq\sum_{j \neq 1} \bigg(\Delta_j \cdot \sum_{t=k+1}^T\exp{-\frac{t\Delta_j^2}{2\Psi_j + \Delta_j}}  - \frac{k}{2n} \cdot (\Psi_j - \Delta_j)\bigg) \\
    &+ \frac{k}{2} \cdot \left(\frac{1}{n} \cdot \sum_{j \neq 1} (\Psi_j - \Delta_j) - (\Psi-\Delta)\right)\\
    &\leq \sum_{j \neq 1} \bigg(\Delta_j \cdot \sum_{t=k+1}^T\exp{-\frac{t\Delta_j^2}{2\Psi_j + \Delta_j}} - \frac{k}{2n} \cdot (\Psi_j - \Delta_j)\bigg).
\end{align*}
For each $j \neq 1$, an identical derivation to the one in the case of two actions would yield the same regret rate, with $k/n$ in place of $k$. Thus, overall, we obtain
\[
    R_T(\cA_k) \leq \frac{5nT}{k},
\]
which concludes the proof.
\end{proof}
\section{Conclusions}\label{sec:conclusion}

Our work introduces \textit{best-action queries} in the context of online learning. We provide tight minimax regret in both the full feedback model and in the label efficient one. We establish that leveraging a sublinear number of best action queries is enough to improve significantly the regret rates achievable \emph{without} best-action queries. 
%
Promising avenues for future research involve integrating best-action queries with diverse feedback forms, extending beyond full feedback, such as bandit feedback, partial monitoring, and feedback graphs (where, in particular, Lemma~\ref{lem:potential-hedge} does not hold). Moreover, our work only studies the case where queries are \textit{perfect}, i.e., the queried oracle gives the correct identity of the best action at that time step with probability $1$. Imagining a \textit{noisy} oracle that gives the correct identity of the best action only with probability $\nicefrac{1}{n} + \delta$ is also an interesting future direction this work leaves open.

\section*{Acknowledgments}
Federico Fusco, Stefano Leonardi and Matteo Russo are partially supported by the ERC Advanced Grant 788893 AMDROMA ``Algorithmic and Mechanism Design Research in Online Markets''.
Andrea Celli is partially supported by MUR - PRIN 2022 project 2022R45NBB funded by the NextGenerationEU program. Federico Fusco is also partially supported by the FAIR (Future Artificial Intelligence Research) project PE0000013, funded by the NextGenerationEU program within the PNRR-PE-AI scheme (M4C2, investment 1.3, line on Artificial Intelligence), and by the PNRR MUR project IR0000013-SoBigData.it.

\bibliography{references}

\end{document}